\documentclass[10pt,twocolumn,letterpaper]{article}

\usepackage{cvpr}              

\usepackage[dvipsnames]{xcolor}

\definecolor{cvprblue}{rgb}{0.21,0.49,0.74}
\usepackage[pagebackref,breaklinks,colorlinks,citecolor=cvprblue]{hyperref}

\def\paperID{12000} 
\def\confName{CVPR}
\def\confYear{2024}
\title{APSeg: Auto-Prompt Network for Cross-Domain Few-Shot\\Semantic Segmentation}
\author{Weizhao He$^{1\dag}$, Yang Zhang$^{1\dag*}$, Wei Zhuo$^{3*}$, Linlin Shen$^{1,2,3}$, Jiaqi Yang$^{3, 4}$, Songhe Deng$^1$, Liang Sun$^1$\\
    $^1$Computer Vision Institute, School of Computer Science \& Software Engineering, Shenzhen University\\
    $^2$Shenzhen Institute of Artificial Intelligence and Robotics for Society\\
    $^3$National Engineering Laboratory for Big Data System Computing Technology, Shenzhen University\\
    $^4$School of Computer Science, University of Nottingham, China\\
    {\small \{heweizhao, dengsonghe, sunliang\}2022@email.szu.edu.cn}~
    {\small \{yangzhang, weizhuo, llshen\}@szu.edu.cn}~
    {\small jiaqi.yang2@nottingham.edu.cn}
}

\usepackage{multirow}
\usepackage{stfloats}
\usepackage{tabularx}
\usepackage{indentfirst}
\usepackage{footmisc}
\usepackage[accsupp]{axessibility}
\usepackage[resetlabels]{multibib}
\newcites{Extra}{References}
\begin{document}
\maketitle
\def\thefootnote{$^{\dag}$}\footnotetext{Equal Contribution: Weizhao He and Yang Zhang}
\def\thefootnote{$^{*}$}\footnotetext{Corresponding Author: Yang Zhang and Wei Zhuo}
\begin{abstract}
Few-shot semantic segmentation (FSS) endeavors to segment unseen classes with only a few labeled samples. Current FSS methods are commonly built on the assumption that their training and application scenarios share similar domains, and their performances degrade significantly while applied to a distinct domain. To this end, we propose to leverage the cutting-edge foundation model, the Segment Anything Model (SAM), for generalization enhancement. The SAM however performs unsatisfactorily on domains that are distinct from its training data, which primarily comprise natural scene images, and it does not support automatic segmentation of specific semantics due to its interactive prompting mechanism. In our work, we introduce APSeg, a novel auto-prompt network for cross-domain few-shot semantic segmentation (CD-FSS), which is designed to be auto-prompted for guiding cross-domain segmentation. Specifically, we propose a Dual Prototype Anchor Transformation (DPAT) module that fuses pseudo query prototypes extracted based on cycle-consistency with support prototypes, allowing features to be transformed into a more stable domain-agnostic space. Additionally, a Meta Prompt Generator (MPG) module is introduced to automatically generate prompt embeddings, eliminating the need for manual visual prompts. We build an efficient model which can be applied directly to target domains without fine-tuning. Extensive experiments on four cross-domain datasets show that our model outperforms the state-of-the-art CD-FSS method by 5.24\% and 3.10\% in average accuracy on 1-shot and 5-shot settings, respectively.
\end{abstract}
    
\section{Introduction}
\label{sec:intro}
\begin{figure}[htbp]
	\centering
	\includegraphics[width=1\linewidth]{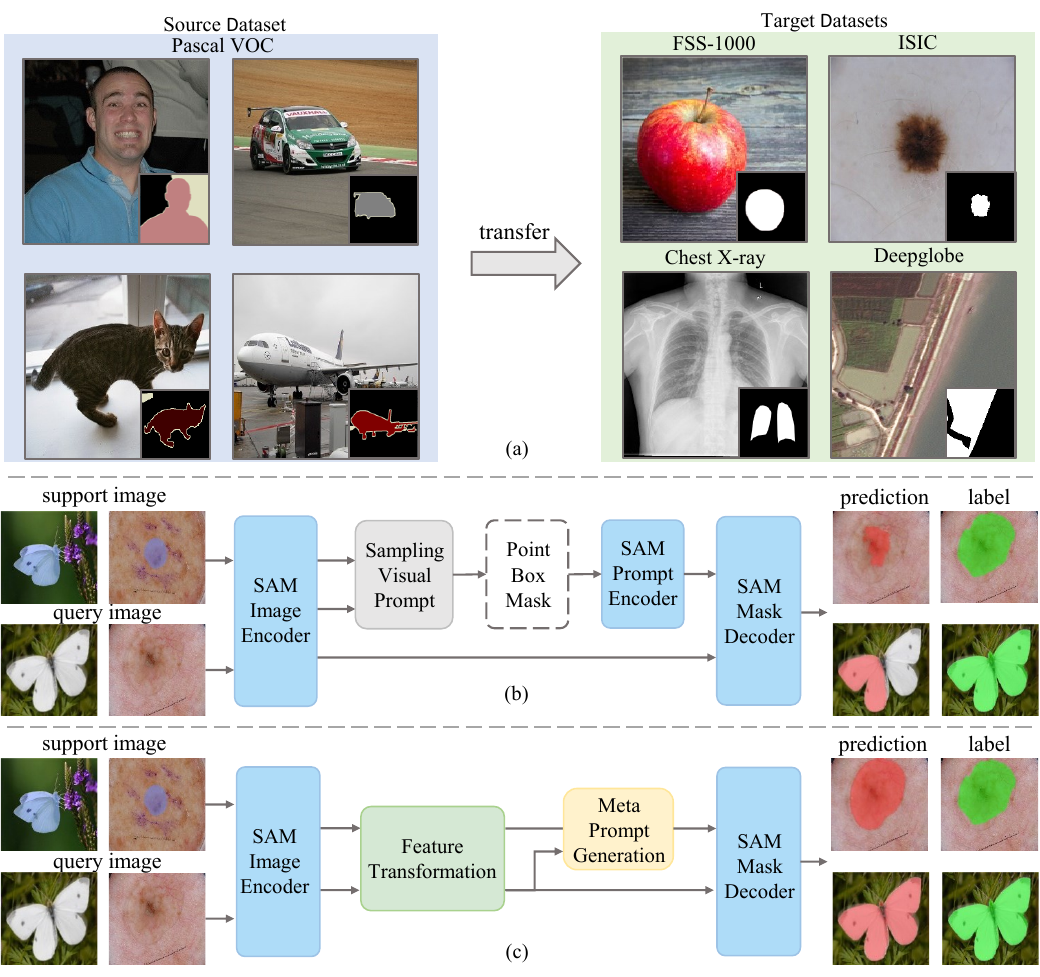}
	\caption{(a) In CD-FSS tasks, training (source) and testing (target) datasets come from different domains, and categories in the testing dataset are unseen during the training phase. (b) The framework of PerSAM \cite{zhang2023personalize}, an existing one-shot segmentation method based on SAM. (C) The framework of our proposed APSeg.}
    \label{fig:intro}
\vspace{-5pt}
\end{figure}
Current deep neural networks \cite{chen2017deeplab, chen2018encoder, long2015fully, xie2021segformer} depend heavily on extensive annotated data to attain satisfactory performance. Data annotation is a time-consuming task that requires significant human resources, especially for dense pixel-wise annotation for segmentation tasks~\cite{chen2018encoder, lang2022learning, li2021adaptive}.
The few-shot semantic segmentation (FSS)~\cite{shaban2017one} is therefore introduced to close this gap, aiming to produce pixel-level predictions for a novel category with only a few labeled samples.
Although existing FSS methods \cite{tian2020prior, peng2023hierarchical, lu2021simpler, lang2022learning} have achieved significant progress, they are commonly built on the assumption that their training and test images are from the same domain. When applied to a different domain, their performance decreases dramatically \cite{wang2022remember, lei2022cross, tavera2022pixel, wang2022cross}. The cross-domain generalizability is thus significant and necessary. FSS models typically require a large amount of base data for training. However, images for some tasks, such as cancer diagnosis and remote-sensing analysis, are scarce and challenging to obtain, making training a powerful model directly on their own data nearly impossible. In our work, we aim to transfer knowledge from easily accessible natural domains to data-scarce domains, as shown in Fig.~\ref{fig:intro}(a). We believe that the transfer ability of general models trained on large-scale natural scene datasets will benefit domains that own only a handful of data. 

To accomplish the cross-domain few-shot semantic segmentation (CD-FSS) task, PATNet \cite{lei2022cross} utilizes support prototypes for computing a transformation matrix, facilitating the conversion of domain-specific features into domain-agnostic ones. In addition, PATNet has a further fine-tuning process using the target domain data during the testing phase.
Built on \cite{lei2022cross}, RestNet \cite{huang2023restnet} introduces a unified attention module to enhance query and support features prior to transformation. Residual connections are integrated to fuse features before and after the transformation, preserving important information from the original space. Furthermore, RestNet achieves better segmentation results by predicting twice. 
These works, however, are inefficient in application due to an additional finetuning process or double predictions. In addition, their backbone pretrained on ImageNet \cite{russakovsky2015imagenet} limits their performance. 

To this end, we attempt to take advantage of recent achievements on the foundation model, the Segment Anything Model (SAM) \cite{kirillov2023segment} to assist CD-FSS. 
With more than one billion masks under its training, SAM exhibits strong feature extraction and generalization abilities. However, recent studies \cite{mazurowski2023segment, roy2023sam, chen2023sam, chen2024rsprompter} have reported that applying SAM directly to new domains often yields subpar performance on zero-shot segmentation tasks, particularly when the data distribution significantly differs from the natural domain data used in SAM training.
In addition, its interactive framework necessitates manual visual prompts, such as points or boxes, for precise segmentation, which restricts its capability for full automation.
Furthermore, some recent works \cite{yue2023surgicalsam, chen2024rsprompter, li2023auto} demonstrate SAM is sensitive to manual visual prompts. Even slight deviations in provided prompts can remarkedly affect segmentation accuracy. As shown in Fig.~\ref{fig:intro}(b), PerSAM presents unsatisfactory performance due to its inability to extract high-quality visual prompts.

In our work, we propose an auto-prompt network for cross-domain few-shot semantic segmentation (APSeg), which builds a novel end-to-end framework that efficiently adapts SAM to CD-FSS tasks for accurate segmentation. As shown in Fig.~\ref{fig:intro}(c), the core of our framework is feature transformation and meta prompt generation. In particular, for feature transformation, we propose a Dual Prototype Anchor Transformation (DPAT) module to extract pseudo query prototypes based on cycle-consistency between support and query features. 
By fusing the pseudo query prototypes with the support prototypes, the computation of the transformation matrix can incorporate information from both support and query samples, which facilitates transforming input features into a more resilient domain-agnostic feature space. Combined with DPAT, the potential of SAM in cross-domain scenarios can be unleashed.
In addition, for automatic prompt-embedding generation, we introduce a Meta Prompt Generator (MPG) module via a meta-learning procedure. Rather than relying on manual visual prompts such as point and box prompts, MPG leverages support features to guide the generation of meta prompt embeddings associated with target objects to substitute the output of the SAM's prompt encoder. With MPG, our method is robust and general for automatic segmentation. Our main contributions are summarized as follows:
\begin{itemize}
    \item We propose a novel model that integrates a dual prototype anchor transformation (DPAT) module and a meta prompt generator (MPG) module for efficiently adapting SAM to CD-FSS tasks.
    \item The DPAT module is proposed for cross-domain feature transformation, which integrates support prototypes and pseudo query prototypes and transforms input features into a stable domain-agnostic space. 
    \item The MPG module is introduced to generate prompt embeddings through meta-learning to establish a fully automatic framework for segmentation.
    \item Extensive experiments on four cross-domain datasets demonstrate that our model outperforms the state-of-the-art CD-FSS method by 5.24\%  and 3.10\% in average accuracy on 1-shot and 5-shot settings, respectively. Especially, we attain 17.49\% (1-shot) and 14.30\% (5-shot) improvements on the Chest X-ray dataset.
\end{itemize}

In our method, the parameters of SAM’s image encoder and mask decoder are frozen, and only a few parameters are trainable. Notably, our trained model can directly achieve promising results when applied to target domains without fine-tuning or 
multiple rounds of inference.

\section{Related Work}
\label{sec:related_work}
\begin{figure*}[htbp]
    \centering
    \includegraphics[width=0.9\textwidth]{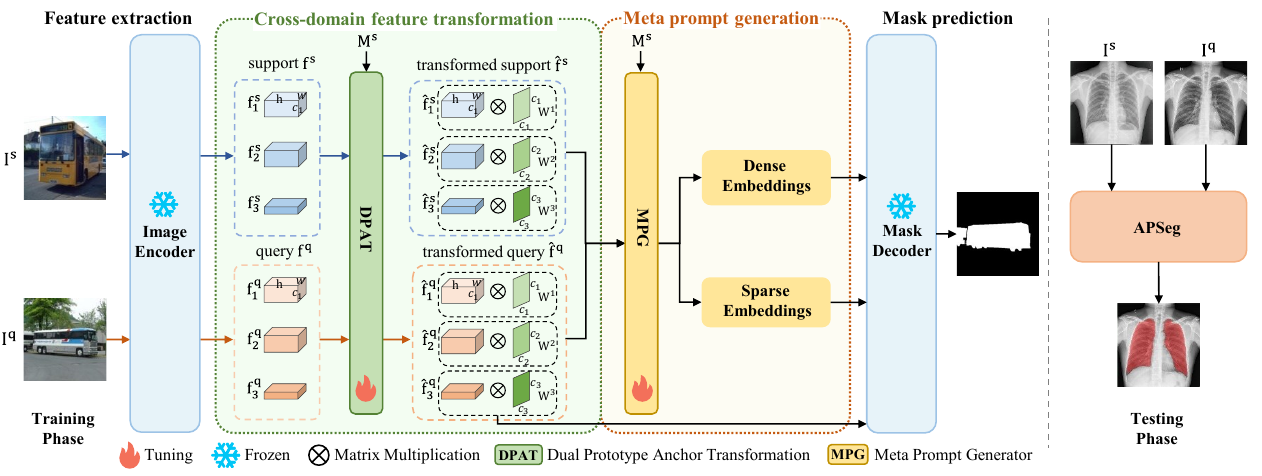}
    \caption{The overall architecture of our proposed APSeg in a 1-way 1-shot example. After obtaining the multi-layer features of support and query images, DPAT is employed to transform the domain-specific features into domain-agnostic ones through linear transformation. Then, the transformed features are passed into MPG to generate prompt embeddings. At last, the mask decoder takes the prompt embeddings and the transformed high-level query feature as input to make a prediction for the query image. In the testing phase, the trained model is directly applied to complete meta-testing in the target domain.}
    \label{fig:overview}
\vspace{-10pt}
\end{figure*}

\subsection{Few-shot Segmentation}
The goal of few-shot semantic segmentation (FSS) is to segment new classes with a few annotated examples.
Current FSS methods are commonly based on meta-learning, which can be largely grouped into two types: prototype-based methods d\cite{wang2019panet, yang2021mining, zhang2020sg, cao2022prototype} and matching-based methods \cite{lang2022learning, wu2021learning, zhang2021self, peng2023hierarchical}.
Motivated by PrototypicalNet \cite{snell2017prototypical} for few-shot learning, the prevalent FSS models utilize prototypes for specific-class representation. Recent works \cite{li2021adaptive, zhang2019pyramid} point out that a single prototype has a limitation to cover all regions of an object, especially for pixel-wise dense segmentation tasks. To alleviate this problem, methods of \cite{li2021adaptive, zhang2019pyramid} use expectation-maximization and cluster algorithms to generate multiple prototypes for different parts of the objects.
Compared with prototype-based methods, matching-based ones \cite{tian2020prior, zhang2021few, min2021hypercorrelation} are designed to extract dense correspondences between query images and support annotations, harnessing pixel-level features to augment the support context with more intricate details. These methods, however, only focus on segmenting new categories from the same domain and fail to generalize unseen domains.

\subsection{Cross-domain Few-shot Segmentation}
In contrast to the traditional FSS setting, CD-FSS necessitates that models refrain from accessing target data during the training process. Furthermore, the data distribution and labeling space in the test phase differ from those in the training phase. This is a more realistic setting.
PATNet \cite{lei2022cross} proposes a feature transformation module which aims to convert domain-specific features into domain-agnostic features. In addition, the target domain data are desired to be utilized to fine-tune the model during the testing phase. Notably, PATNet outperforms current FSS methods on its established benchmark. RestNet \cite{huang2023restnet} utilizes a lightweight attention module to enhance pre-transformation features and merge post-transformation features through residual connections to maintain the key information in the original domain. Meanwhile, a mask prediction strategy is introduced to mitigate the issue of overfitting to support samples in FSS and facilitates the model in a gradual acquisition of cross-domain segmentation capabilities. However, existing methods still utilize classical classification models \cite{simonyan2014very, he2016deep} as feature extractors with limited feature extraction capabilities compared to existing visual foundation models \cite{kirillov2023segment, oquab2023dinov2, radford2021learning}. 
The performance of cross-domain segmentation is limited. Moreover, either additional training or multiple inferences is required when predicting masks, which makes the inference process complex and inefficient.

\subsection{Segment Anything Model}
The Segment Anything Model (SAM) \cite{kirillov2023segment}, pretrained on massive amounts of labeled data, first introduced a foundation model for image segmentation. SAM relies on explicit points and bounding boxes at precise locations for accurate segmentation \cite{cheng2023sam, roy2023sam}. Therefore, extensive manual guidance or a specialist detector is often required, leading to a complex multi-stage pipeline \cite{wang2023sam}. SAM cannot achieve automatic segmentation for specific semantics. To address this, PerSAM \cite{zhang2023personalize} proposes automatic sampling of visual prompts, and some other methods suggest directly generating prompt embeddings \cite{yue2023surgicalsam, chen2024rsprompter}. Inspired by these works, we propose an automatic prompting method in a meta-learning manner to adapt SAM to CD-FSS. 
\section{Method}
\label{sec:method}
\subsection{Problem Definition}
For CD-FSS,  a source domain $(X_s, Y_s)$ and a target domain $(X_t, Y_t)$ exist, where the input distribution of the source domain and target domain are different, and their label space has no intersection as well, i.e., $X_s \neq X_t$ and $Y_s \cap Y_t = \emptyset$. Here $X$ denotes input distribution, and $Y$ denotes the label space. In our work, we train and test our model in a meta-learning episodic manner following~\cite{lei2022cross}, and our model is only trained on the source domain and has no access to the target data. Each episode data consists of a support set $S$ and a query set $Q$ with a specific category. The support set $S = {(\mathrm{I}_i^{\mathrm{s}}, \mathrm{M}_i^{\mathrm{s}})}^K_{i = 1}$ contains $K$ image-mask pairs, where $\mathrm{I}_i^{\mathrm{s}}$ denotes the $i$-th support image and $\mathrm{M}_i^\mathrm{s}$ denotes the corresponding binary mask. 
Similarly, the query set is defined as $Q = ({\mathrm{I}_i^{\mathrm{q}}, \mathrm{M}_i^\mathrm{q})}^K_{i = 1}$. 
To train our model, support sets and images from query sets are used as model inputs to predict the query masks.
To assess the trained model's performance, we test it on a support set and a query set from the target domain.

\begin{figure}
\centering
    \includegraphics[width=1\linewidth]{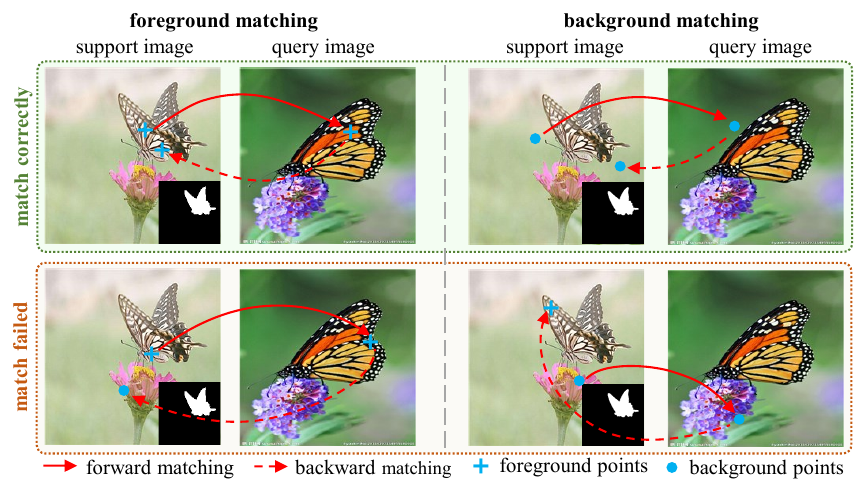}
    \caption{A visual example of a support-query pair to perform cycle-consistent selection (CCS).}
    \label{fig:dpat-illu}
\vspace{-10pt}
\end{figure}

\subsection{Method Overview}
Our target is to train a general model on natural domains with rich annotations and transfer the knowledge to target domains with limited labeled data.
As illustrated in Fig.~\ref{fig:overview}, our proposed APSeg consists of two key modules: the Dual Prototype Anchor Transformation (DPAT) module and the Meta Prompt Generator (MPG) module. 
Specifically, given the support image $\mathrm{I}^\mathrm{s}$ and query image $\mathrm{I}^\mathrm{q}$, the SAM image encoder is used to extract multi-level features from different layers.
DPAT module is then employed to map support and query features to a new domain-agnostic space, facilitating the rapid adaptation of subsequent modules for previously unseen domains.
Next, we introduce the MPG module whose task is to generate sparse and dense prompt embeddings through the meta-learning procedure for the SAM's mask decoder by utilizing the transformed features.
At last, the generated prompt embeddings and the transformed high-level query features are passed into the mask decoder for target mask prediction.

\subsection{Dual Prototype Anchor Transformation }
\begin{figure}
    \centering
    \includegraphics[width=1\linewidth]{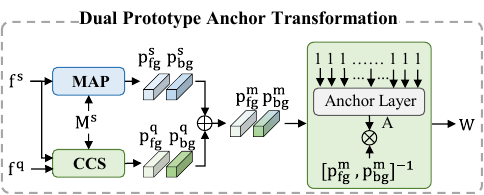}
    \caption{The specific implementation of the Dual Prototype Anchor Transformation module. Pseudo foreground and background prototypes of query are extracted through cycle-consistent selection (CCS). The extracted pseudo prototypes are then fused with support prototypes to calculate the transformation matrix $\mathbf{W}$ with an anchor layer.}
    \label{fig:dpat}
\vspace{-10pt}
\end{figure}

To map support and query features into a new domain-agnostic space, we introduce a Dual Prototype Anchor Transformation (DPAT) module, as shown in Fig.~\ref{fig:dpat}. Previous method~\cite{lei2022cross} solely relies on a set of support prototypes and anchor layers to compute a transformation matrix. However, support prototypes cannot well represent complete information of a category due to intra-class variance. Therefore, we propose to enhance the support prototype set with query prototypes for better feature transformation.
Specifically, our proposed DPAT consists of two procedures: dual prototype enhancement and cross-domain feature transformation. Inspired by \cite{zhang2021few}, we propose cycle-consistent selection (CCS) to extract both the query foreground and background prototypes in the absence of query masks, which is used to enhance the support prototypes. Based on these enhanced prototypes which represent a category and its surroundings, an effective transformation matrix can be computed with a learnable domain-agnostic module. The transformation matrix is then applied to support and query features for cross-domain feature transformation.

\paragraph{Dual Prototype Enhancement} Representative prototypes are significant for our cross-domain transformation. To this end, we build a cycling-examine procedure that reasons on both the query foregrounds and backgrounds to augment support prototypes. The process is shown in Fig.~\ref{fig:dpat-illu}. We conduct forward matching to attain query features that have the highest similarity with the support foregrounds. We then use these identified forward-matched query features to re-locate corresponding support features reversely. If the located support features through reverse matching fall within support foregrounds, the identified query features will be averaged and used to derive the foreground prototypes. The background prototypes are obtained with the same process. Finally, support prototypes and query prototypes are fused by addition.

Specifically, given a support image $\mathrm{I}^\mathrm{s}$ and a query image $\mathrm{I}^\mathrm{q}$, we initially employ a shared-weight image encoder to extract their multi-layer feature maps $\left\{ \mathrm{f}^\mathrm{s}_l \right\}^3_{l=1}$ and $\left\{ \mathrm{f}^\mathrm{q}_l \right\} ^3_{l=1}$ respectively, where ${\mathrm{f}^\mathrm{s}_l}$, ${\mathrm{f}^\mathrm{q}_l}\in\mathbb{R}^{c_l\times h \times w}$. $c_l, h, w$ are the channel dimension, height, and width of the feature map.
DPAT takes support features$\left\{ \mathrm{f}^\mathrm{s}_l \right\}^3_{l=1}$, query features$\left\{ \mathrm{f}^\mathrm{q}_l \right\} ^3_{l=1}$ and support mask $\mathrm{M}^\mathrm{s}$ as input. To simplify the notations, $\mathrm{f}^\mathrm{s}$ and $\mathrm{f}^\mathrm{q}$ are used to represent any one of $\left\{ \mathrm{f}^\mathrm{s}_l \right\}^3_{l=1}$ and $\left\{ \mathrm{f}^\mathrm{q}_l \right\} ^3_{l=1}$. We then employ masked average pooling (MAP) on the support feature to obtain the foreground and background prototype, denoted as $\mathbf{p}^{\mathrm{s}}_{\mathrm{fg}}$ and $\mathbf{p}^{\mathrm{s}}_{\mathrm{bg}}$. 
By concatenating $\mathbf{p}^{\mathrm{s}}_{\mathrm{fg}}$ and $\mathbf{p}^{\mathrm{s}}_{\mathrm{bg}}$, we get the support prototype matrix $\mathbf{P}^{\mathrm{s}}=\left[\mathbf{p}^{\mathrm{s}}_{\mathrm{fg}}, \mathbf{p}^{\mathrm{s}}_{\mathrm{bg}}\right]$.
Considering that the query mask cannot be accessed during training, we extract pseudo query prototypes through CCS. First, the support foreground feature is obtained by multiplying the support mask $\mathrm{M}^\mathrm{s}$ with the support feature $\mathrm{f}^\mathrm{s}$.
Next, the similarity between the support foreground feature and the query feature is calculated. For each element of the support foreground feature, we search for the element in the query feature map with the highest similarity score, and acquire the matched position set $i^{\mathrm{s} \rightarrow \mathrm{q}}$ from support to query as follows,
\begin{equation}
i^{\mathrm{s} \rightarrow \mathrm{q}}=\underset{i\in\{0,1,...,h\times w -1\}}{\operatorname*{\arg\max}}(\mathrm{sim}(\mathrm{f}^\mathrm{s} \odot \mathrm{M}^\mathrm{s}, \mathrm{f}^\mathrm{q}_i))
\end{equation}
where $\mathrm{sim}(\cdot,\cdot)$ is a cosine function. Based on $i^{\mathrm{s} \rightarrow \mathrm{q}}$, the matched query feature $\mathrm{f}^\mathrm{q}_{i^{\mathrm{s} \rightarrow \mathrm{q}}}$ can be extracted.
Similarly, the matched positions $j^{\mathrm{s} \leftarrow \mathrm{q}}$ from query to support can obtained as below,
\begin{equation}
\mathrm{f}^\mathrm{q}_{i^{\mathrm{s} \rightarrow \mathrm{q}}}=\{\mathrm{f}^\mathrm{q}[i]:i\in i^{\mathrm{s} \rightarrow \mathrm{q}}\}
\end{equation}
\begin{equation}
j^{\mathrm{s} \leftarrow \mathrm{q}}=\underset{j\in\{0,1,...,h\times w -1\}}{\operatorname*{\arg\max}}(\mathrm{sim}(\mathrm{f}^\mathrm{q}_{i^{\mathrm{s} \rightarrow \mathrm{q}}}, \mathrm{f}^\mathrm{s}_j))
\end{equation}
where $j^{\mathrm{s} \leftarrow \mathrm{q}}$ is the positions on support that have the most similar features with the corresponding reference query features $\mathrm{f}^\mathrm{q}_{i^{\mathrm{s} \rightarrow \mathrm{q}}}$. 
If the position in the matched position set $j^{\mathrm{s} \leftarrow \mathrm{q}}$ does not fall in the support mask $\mathrm{M}^\mathrm{s}$, we filter out the position from $i^{\mathrm{s} \rightarrow \mathrm{q}}$ and obtain the final matched position set $i^{'\mathrm{s} \rightarrow \mathrm{q}}$.
According to the set $i^{'\mathrm{s} \rightarrow \mathrm{q}}$, the corresponding features are extracted from $\mathrm{f}^\mathrm{q}$ and averaged to obtain the pseudo foreground query prototype $\mathbf{{p}}^{\mathrm{q}}_\mathrm{fg}$.
The pseudo background query prototype $\mathbf{{p}}^{\mathrm{q}}_\mathrm{bg}$ can be obtained through a similar process, resulting in the query prototype matrix $\mathbf{{P}}^{\mathrm{q}}=\left[\mathbf{{p}}^{\mathrm{q}}_\mathrm{fg}, \mathbf{{p}}^{\mathrm{q}}_\mathrm{bg}\right]$.
Finally, a mixed prototype matrix can be obtained by $\mathbf{P}^\mathrm{m} = \mathbf{{P}}^{\mathrm{s}}$ + $\mathbf{{P}}^{\mathrm{q}}$.

\paragraph{Cross-domain Feature Transformation}
Features of the same class yield similar results when they are transformed in the same way. Support features and query features are transformed into a domain-agnostic space using the same transformation matrix $\mathbf{W}$ to avoid the detrimental impact caused by domain shift.
Given the weight matrix of an anchor layer $\mathbf{A}$, the definition of a transformation matrix is as follows,
\begin{equation}
\mathbf{W}\mathbf{P}^\mathrm{m}=\mathbf{A}
\end{equation}
where $\mathbf{P}^\mathrm{m} = \left[\frac{\mathbf{p}^{\mathrm{m}}_{\mathrm{fg}}}{\|\mathbf{p}^{\mathrm{m}}_{\mathrm{fg}}\|},\frac{\mathbf{p}^{\mathrm{m}}_{\mathrm{bg}}}{\|\mathbf{p}^{\mathrm{m}}_{\mathrm{bg}}\|}\right]$, $\mathbf{A} = \left[\frac{\mathbf{a}_{\mathrm{fg}}}{\|\mathbf{a}_{\mathrm{fg}}\|},\frac{\mathbf{a}_{\mathrm{bg}}}{\|\mathbf{a}_{\mathrm{bg}}\|}\right]$ and $\mathbf{a}$ is the anchor vector which is independent of the input.
\begin{figure}
\centering
    \includegraphics[width=0.9\linewidth]{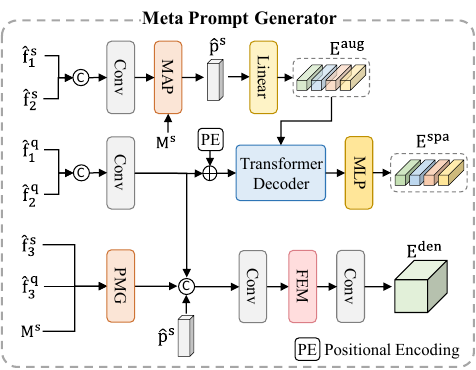}
    \caption{The specific implementation of the Meta Prompt Generator module. PE indicates a positional encoding.}
    \label{fig:mpg}
\vspace{-10pt}
\end{figure}
Different from previous work \cite{lei2022cross}, we leverage the mixed prototype matrix that incorporates both support and query feature information during the computation process. Since the prototype $\mathbf{P}^\mathrm{m}$ is a non-square matrix, the generalized inverse \cite{ben2003generalized} of $\mathbf{P}^\mathrm{m}$ is calculated with $\mathbf{P}^{\mathrm{m+}}=\{{\mathbf{P}^\mathrm{m}}^\mathbf{T} \mathbf{P}^{\mathrm{m}}\}^{-1}{\mathbf{P}^{\mathrm{m}}}^\mathbf{T}$. 
Therefore, the transformation matrix is calculated as
$\mathbf{W}=\mathbf{A}\mathbf{P}^{\mathrm{m+}}$. In our work, 
We have two different anchor layers for mid-level features $\{\mathrm{f}^\mathrm{s}_1, \mathrm{f}^\mathrm{s}_2, \mathrm{f}^\mathrm{q}_1, \mathrm{f}^\mathrm{q}_2 \}$ and high-level features $\{\mathrm{f}^\mathrm{s}_3, \mathrm{f}^\mathrm{q}_3\}$. Finally, we can efficiently map support and query features to a stable, domain-agnostic space by multiplying them with $\mathbf{W}$. 

Objects and things of even the same class can differ in shape and appearance. Due to limited samples of supports, it is intrinsically challenging to represent all the variance within objects and things of a class. Through the double-check procedure for both foreground and background regions, our proposed DPAT can effectively mitigate the challenges aroused by intra-class variances and generate a more stable transformation matrix for cross-domain feature transformation. 

\subsection{Meta Prompt Generator}
To construct an end-to-end fully automated SAM-based segmentation framework for CD-FSS, we utilize the features of support-query pairs to directly generate prompt embeddings. In particular, a meta prompt generator (MPG) module is designed to obtain both sparse and dense embeddings simultaneously, as shown in Fig.~\ref{fig:mpg}. Different from~\cite{cao2022prototype} that uses single support embeddings without alignment, our new pipeline extends to leverage multiple support embeddings and integrates feature alignment. Our design comprehensively takes account of intra-class variance and the multiple prompts coming along with the support embeddings enhance the segmentation.
For clarity, we refer to the process of generating sparse embeddings and dense embeddings as sparse path and dense path respectively. In this way, our method eliminates the need for external manual visual prompts, such as points or boxes.
 
\textbf{Sparse path.} In this process, the query features and several support embeddings augmented from the support prototypes are utilized to generate sparse embeddings through a transformer decoder~\cite{carion2020end}, which then replace the original sparse embeddings in SAM. First, we concatenate the transformed mid-level support features $\hat{\mathrm{f}}^\mathrm{s}_1\in\mathbb{R}^{c_1 \times h \times w}$ and $\hat{\mathrm{f}}^\mathrm{s}_2\in\mathbb{R}^{c_2 \times h \times w}$ along the channel dimension, and then perform dimension reduction through a convolution layer to obtain $\hat{\mathrm{f}}^\mathrm{s}\in\mathbb{R}^{c_{r} \times h \times w}$. In the same way, we can also get $\hat{\mathrm{f}}^\mathrm{q}\in\mathbb{R}^{c_{r} \times h \times w}$,
\begin{equation}
\begin{aligned}
\hat{\mathrm{f}}^\mathrm{s}=\mathcal{F}_{conv}(\hat{\mathrm{f}}^\mathrm{s}_1\oplus \hat{\mathrm{f}}^\mathrm{s}_2) \\
\hat{\mathrm{f}}^\mathrm{q}=\mathcal{F}_{conv}(\hat{\mathrm{f}}^\mathrm{q}_1\oplus \hat{\mathrm{f}}^\mathrm{q}_2)
\end{aligned}
\end{equation}
where $\mathcal{F}_{conv}$ means performing a $1\times 1$ convolution followed by a ReLU activation function and $\oplus$ denotes the concatenation operation in channel dimension.

Next, we take $\hat{\mathrm{f}}^\mathrm{s}$ and $\mathrm{M}^\mathrm{s}$ as input and apply MAP to obtain the foreground class prototypes $\mathbf{\hat{p}}^s\in\mathbb{R}^{c_{r}}$. Then, a linear layer maps the $\mathbf{\hat{p}}^s$ to multiple augmented support embeddings $\mathbf{E}^{\mathrm{aug}}\in\mathbb{R}^{k \times c_{r}}$,
\begin{equation}
\mathbf{E}^{\mathrm{aug}}=\mathcal{F}_{linear}(\mathbf{\hat{p}}^\mathrm{s})
\end{equation}
where $k$ denotes the number of the augmented support embeddings. Here we generate several embeddings instead of only a single embedding for simulating multiple point prompts.

To supplement positional information, learnable position encodings are applied to $\mathbf{E}^{\mathrm{aug}}$ and fixed sine-cosine positional encodings are applied to $\mathrm{\hat{f}}^\mathrm{q}$. We then input them into the transformer decoder and its output is further processed by a two-layer MLP to increase the channel dimensions, which yields $\hat{\mathbf{E}}^{\mathrm{aug}}\in\mathbb{R}^{k \times c_{o}}$. 
\begin{equation}
\mathbf{\hat{E}}^{\mathrm{aug}}=\mathcal{F}_{mlp}(\mathcal{F}_{trans}(\mathbf{E}^{\mathrm{aug}}, \mathrm{\hat{f}}^\mathrm{q}))
\end{equation}

To align the generated sparse embeddings with those produced by SAM’s prompt encoder, the sine function is employed to generate the final sparse embeddings $\mathbf{E}^{\mathrm{spa}}\in\mathbb{R}^{k \times c_{o}}$ following~\cite{chen2024rsprompter}.

\begin{equation}
\mathbf{E}^{\mathrm{spa}}= \mathbf{\hat{E}}^{\mathrm{aug}} + \mathcal{F}_{sine}(\mathbf{\hat{E}}^{\mathrm{aug}})
\end{equation}

\textbf{Dense path.} In this process, the dense embeddings are modulated by query features and support prototypes.~$\hat{\mathrm{f}}^\mathrm{s}_3$,~$\hat{\mathrm{f}}^\mathrm{q}_3$~and~$\mathrm{M}^\mathrm{s}$~are first passed to prior mask generation (PMG) module~\cite{tian2020prior} to generate a prior mask $\mathrm{M}^{\mathrm{pr}}\in\mathbb{R}^{1\times h \times w}$.
After concatenating $\mathbf{\hat{p}}^{\mathrm{s}}$, $\hat{\mathrm{f}}^\mathrm{q}$ and $\mathrm{M}^{\mathrm{pr}}$, we perform a $1\times 1$ convolution for dimension reduction to obtain $\mathrm{\hat{f}}^{\mathrm{pr}}\in\mathbb{R}^{c_r \times h \times w}$. The output is then passed into the feature enhancement (FEM) module~\cite{tian2020prior} to get $\hat{\mathrm{f}}^{\mathrm{fem}}$.
$\hat{\mathrm{f}}^{\mathrm{fem}}$ is further processed by a $1\times1$ convolution layer to increase the channel dimensions to obtain $\mathbf{E}^{\mathrm{den}}\in\mathbb{R}^{c_o \times h \times w}$.

\subsection{Training Loss}
In the training of APSeg, we employ a Dice loss function, computed between the predicted mask $\hat{\mathrm{M}}$ and the corresponding ground truth query mask $\mathrm{M}^\mathrm{q}$. The loss function, denoted as $\mathcal{L}$, is expressed as:

\begin{equation}\label{loss}
\mathcal{L} = \frac{1}{n} \sum_{i=1}^{n} \operatorname{DICE}\left(\mathcal{I}(\hat{\mathrm{M}}), \mathrm{M}^\mathrm{q}\right)
\end{equation}

Here, $n$ represents the total number of training episodes in each batch, and DICE signifies the Dice loss function. The function $\mathcal{I}$ serves as an interpolation function, ensuring that $\hat{\mathrm{M}}$ shares the same spatial size as $\mathrm{M}^\mathrm{q}$.

\begin{figure*}[htbp]
    \centering
    \includegraphics[width=0.85\linewidth]{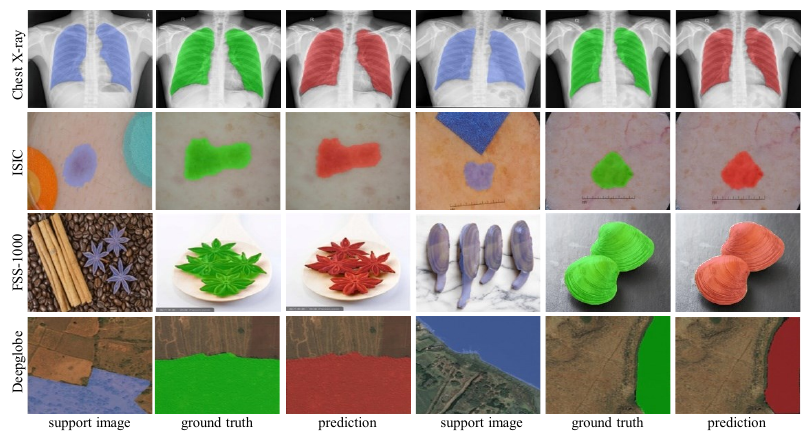}
    \caption{Qualitative results of APSeg in 1-way 1-shot segmentation on CD-FSS. Note that the model is trained on PASCAL VOC. Support labels are overlaid in blue. The prediction and ground truth of query images are in green and red respectively. Best viewed in color.}
    \label{fig:result1}
\end{figure*}

\section{Experiment}
\label{sec:experiment}
\begin{table*}[htbp]
\centering
\setlength{\tabcolsep}{4mm}{
    \scalebox{0.73}{
        \begin{tabular}{c|c|c|c|c|c|c|c|c|c|c|c} 
        \toprule
        \multirow{2}{*}{Methods} & \multirow{2}{*}{Backbone} & \multicolumn{2}{c|}{Chest X-ray} & \multicolumn{2}{c|}{ISIC}  & \multicolumn{2}{c|}{FSS-1000} & \multicolumn{2}{c|}{Deepglobe} & \multicolumn{2}{c}{Average}  \\ 
        \cline{3-12}
                                 &                          & 1-shot & 5-shot           & 1-shot & 5-shot                  & 1-shot & 5-shot              & 1-shot & 5-shot       & 1-shot & 5-shot       \\ 
        \hline
        \multicolumn{12}{c}{Few-shot Semantic Segmentation Methods}                                                                                                                                \\ 
        \hline
        AMP \cite{siam2019amp}                 & VGG-16           & 51.23   & 53.04                    
        & 28.42   & 30.41        
        & 57.18   & 59.24        
        & 37.61   & 40.61      
        & 43.61   & 45.83           
        \\
        PGNet \cite{zhang2019pyramid}          & Res-50           & 33.95   & 27.96                    
        & 21.86   & 21.25         
        & 62.42   & 62.74       
        & 10.73   & 12.36      
        & 32.24   & 31.08           
        \\
        PANet \cite{wang2019panet}             & Res-50           & 57.75   & 69.31                    
        & 25.29   & 33.99       
        & 69.15   & 71.68          
        & 36.55   & \textbf{45.43}      
        & 47.19   & 55.10           
        \\
        CaNet \cite{zhang2019canet}            & Res-50           & 28.35   & 28.62                  
        & 25.16   & 28.22       
        & 70.67   & 72.03         
        & 22.32   & 23.07      
        & 36.63   & 37.99          
        \\
        RPMMs \cite{yang2020prototype}         & Res-50           & 30.11   & 30.82                
        & 18.02   & 20.04     
        & 65.12   & 67.06        
        & 12.99   & 13.47   
        & 31.56   & 32.85 
        \\
        PFENet \cite{tian2020prior}            & Res-50           & 27.22   & 27.57                 
        & 23.50   & 23.83      
        & 70.87   & 70.52       
        & 16.88   & 18.01    
        & 34.62   & 34.98           
        \\
        RePRI \cite{boudiaf2021few}            & Res-50           & 65.08   & 65.48              
        & 23.27   & 26.23   
        & 70.96   & 74.23      
        & 25.03   & 27.41     
        & 46.09   & 48.34        
        \\
        HSNet \cite{min2021hypercorrelation}   & Res-50           & 51.88   & 54.36             
        & 31.20   & 35.10    
        & 77.53   & 80.99     
        & 29.65   & 35.08    
        & 47.57   & 51.38          
        \\ 
        PerSAM \cite{zhang2023personalize}             & ViT-base         & 29.95   & 30.05              
        & 23.27   & 25.33         
        & 60.92   & 66.53  
        & 36.08   & 40.65     
        & 37.56   & 40.64 
        \\
        \hline
        \multicolumn{12}{c}{Cross-domain Few-shot Semantic Segmentation Methods}                                                                                                                   \\ 
        \hline
        PATNet \cite{lei2022cross}             & Res-50           
        & 66.61   & 70.20     
        & 41.16   & 53.58    
        & 78.59   & 81.23     
        & \textbf{37.89}   & 42.97   
        & 56.06   & 61.99       
        \\
        PATNet$^{\ddagger}$ \cite{lei2022cross}             & ViT-base           
        & 76.43   & -     
        & 44.25   & -    
        & 72.03   & -     
        & 22.37   & -   
        & 53.77   & -       
        \\
        RestNet \cite{huang2023restnet}             & Res-50           
        & 70.43   & 73.69     
        & 42.25   & 51.10    
        & \textbf{81.53}   & \textbf{84.89}     
        & -   & -   
        & -   & -       
        \\
        
        \textbf{APSeg(ours)}                  & ViT-base         & \textbf{84.10} 	& \textbf{84.50}                     & \textbf{45.43}    & \textbf{53.98}        
        & 79.71   & 81.90
        & 35.94   & 39.98         
        & \textbf{61.30}    & \textbf{65.09}  
        \\
        \bottomrule
        \end{tabular}
        }
}
\caption{Comparison with previous FSS and CD-FSS methods under 1-way 1-shot and 1-way 5-shot settings. All the methods are trained on the source dataset and tested on the CD-FSS benchmark. The method marked with $\ddagger$ is implemented by ourselves.}
\label{table1}
\vspace{-10pt}
\end{table*}

\noindent\textbf{Datasets.} Following the previous approach~\cite{lei2022cross}, we use PASCAL VOC 2012 \cite{everingham2010pascal} with SBD \cite{hariharan2011semantic} augmentation as training dataset and then evaluate the trained model on Chest X-ray \cite{candemir2013lung, jaeger2013automatic}, ISIC \cite{codella2019skin}, FSS-1000 \cite{li2020fss} and Deepglobe \cite{demir2018deepglobe} respectively.\par

\noindent\textbf{Metric and Evaluation.} We use the mean intersection-over-union (mIoU) as the evaluation metric, which is the same as the previous method. We take the mean-IoU of 5 runs \cite{min2021hypercorrelation} with different random seeds for each test. For all datasets except FSS-1000, each run has 1200 tasks. Every run of FSS-1000 has 2400 tasks. \par

\noindent\textbf{Implementation Details.} 
In our experiments, we employ the base version of the SAM and keep it frozen during training. To be consistent with the original input to SAM, we set spatial sizes of both support and query images to $1024 \times 1024$.
For the SAM image encoder, we utilize feature maps derived from the mid-level features output by the 5\textsuperscript{th} and 8\textsuperscript{th} transformer blocks, in addition to the high-level features obtained from the final output of the image encoder.
Concerning the DPAT module, we create two anchor layers dedicated to mid-level and high-level features, each configured with channel numbers set to 768 and 256, respectively.
For the MPG module, the number of feature channels after dimension reduction is set to 64. Additionally, the number of sparse embeddings and dense embeddings channels output by MPG is set to 256.
We implement the model in PyTorch and utilize the Adam \cite{kingma2014adam} optimizer with a learning rate of $1\mathrm{e}\mathrm{-}3$.

\subsection{Comparison with State-of-the-Arts}
We compare our method against existing CNN-based and SAM-based approaches for CD-FSS. As shown in Tab.~\ref{table1}, the results demonstrate the superiority of the proposed method in this challenging task. 
Specifically, our approach exhibits improvements of 5.40\% and 3.10\% compared to the PATNet \cite{lei2022cross}, under 1-shot and 5-shot settings, respectively. Moreover, APSeg surpasses the SAM-based method PerSAM \cite{zhang2023personalize} by 23.74\% and 24.45\%, affirming the effectiveness of our approach.
Notably, our method showcases significantly superior performance compared to the current methods when confronted with large domain gaps between the testing and training datasets. Specifically, compared with PATNet, our method achieves improvements of 17.49\% and 14.30\% on the chest X-ray dataset under the 1-shot and 5-shot settings, respectively. Similarly, on the ISIC dataset, improvements of 4.27\% are observed for the 1-shot setting.
Furthermore, our automatic prompting approach significantly surpasses the manual prompting strategy proposed by \cite{zhang2023personalize} in cross-domain performance.
For a fair comparison, we also implement PATNet based on the SAM's image encoder. The model is trained with the same input image size, and test-time fine-tuning is not employed. The results demonstrate that APSeg still maintains a significant superiority.
Qualitative results, illustrated in Fig.~\ref{fig:result1}, validate that our proposed method attains substantial improvements in generalization performance in the presence of large domain gap while maintaining considerable accuracy with minor domain shift. More visualization results are provided in the supplementary materials.

\subsection{Ablation study}
\noindent\textbf{Components analysis.}
We assess the effectiveness of our proposed DPAT module and MPG module by using the 1-shot mIoU averaged on four datasets.
To establish a baseline model, we first remove the DPAT module and the sparse path in the MPG module and then replace the final layer of the dense path with a $1\times 1$ convolution layer for predicting segmentation masks. Tab.~\ref{table2} illustrates the impact of each component on model performance.  Overall, the incorporation of the two components suggested in this paper enhances the baseline by 18.44\%. In the second row, MPG leverages the segmentation capabilities of SAM by autonomously generating semantic-aware prompt embeddings, eliminating the need for manual prompts and improving the baseline by 2.80\%.
Upon combining MPG and DPAT, DPAT unleashes the segmentation capabilities of SAM in cross-domain scenarios. It achieves this by transforming input features into a more stable domain-agnostic feature space, resulting in a significant performance improvement of 15.46\% compared to the second row. More analysis and discussion about DPAT are shown in supplementary materials.

\begin{table}[ht]
\vspace{-5pt}
\centering
\setlength{\tabcolsep}{1mm}{
    \scalebox{0.70}{
        \begin{tabular}{c|c|c|c|c|c}
            \toprule
            Method &Chest X-ray &ISIC &FSS-1000 &Deepglobe &Average \\
            \midrule
            Baseline       &28.80 &44.55 &77.90 &20.19 &42.86   \\
            Baseline + MPG &32.62 &36.67 &79.77 &34.31 &45.84         \\
            Baseline + MPG + DPAT &84.10 &45.43 &79.71 &35.94 &61.30   \\
            \bottomrule
        \end{tabular}
    }
}
\caption{Ablation study on key components of APSeg on CD-FSS. Results are averaged over four datasets for 1-shot.}
\label{table2}
\vspace{-15pt}
\end{table}

\begin{table}[ht]
\vspace{-5pt}
\centering
\setlength{\tabcolsep}{6mm}{
    \scalebox{0.73}{
        \begin{tabular}{c|c|c}
                \toprule
                Sparse & Dense & 1-shot mIoU \\
                \midrule
                 \checkmark &    & 41.52        \\
                  & \checkmark & 44.09        \\
                 \checkmark &\checkmark    & 45.43        \\
                \bottomrule
        \end{tabular}
    }
    
}
\caption{Ablation study on the choice of different types of prompt embeddings in MPG on ISIC.}
\label{table3}
\vspace{-10pt}
\end{table}

\noindent\textbf{Meta Prompt Generator.} Tab.~\ref{table3} shows the impact of the main components in the MPG, namely sparse embeddings and dense embeddings. We show the results for three combination scenarios: using only sparse embeddings, only dense embeddings, and both. Our observation reveals that better performance can be attained when combining sparse and dense embeddings. This emphasizes the significance of leveraging both types of embeddings to harness the segmentation capabilities of SAM in cross-domain scenarios.

\noindent\textbf{The number of feature channels.} After cross-domain feature transformation, we fuse the transformed mid-level features and perform dimension reduction, which can reduce the number of learnable parameters and avoid the trained model overfitting the training dataset. Observation in Tab.~\ref{table4} shows that reducing the number of channels to 64 allows our method to achieve better CD-FSS performance with only a small number of additional learnable parameters.

\begin{table}[ht]
\centering
\setlength{\tabcolsep}{5mm}{
    \scalebox{0.73}{
        \begin{tabular}{c|c|c}
            \toprule
            dim& 1-shot mIoU & \# Learnable Params \\
            \midrule
            64  & 61.30     & 0.7 M    \\
            128   & 60.03  & 2.3 M       \\
            256   & 58.65 & 8.4 M       \\
            \bottomrule
        \end{tabular}
    }
}
\caption{Ablation study on different output feature channels. Results are averaged over four datasets for 1-shot.}
\label{table4}
\vspace{-20pt}
\end{table}

\noindent\textbf{The number of sparse embeddings.} 
Our proposed MPG introduces an automatic generation mechanism for prompt embeddings, replacing SAM's original method of obtaining them through manual prompts fed into the prompt encoder. Employing more manual visual prompts has been shown to enhance SAM's interactive segmentation performance. Therefore, Tab.~\ref{table5} shows the association between the number of sparse embeddings generated by MPG and the performance of cross-domain few-shot segmentation with APSeg. Notably, generating 4 sparse embeddings results in a 0.55\% improvement in 1-shot scenarios. However, as indicated in the third row, continuing to generate more prompt embeddings may lead to a decline in performance.

\begin{table}[ht]
\vspace{-5pt}
\centering
\setlength{\tabcolsep}{6mm}{
    \scalebox{0.73}{
        \begin{tabular}{c|c}
            \toprule
            num &1-shot mIoU \\
            \midrule
            1       &60.75   \\
            4 &61.30         \\
            8 &60.65    \\
            \bottomrule
        \end{tabular}
    }
}
\caption{Ablation study on the different number of sparse embeddings. Results are averaged over four datasets for 1-shot.}
\label{table5}
\vspace{-15pt}
\end{table}
\vspace{-2pt}

\section{Conclusion}
\vspace{-2pt}
In this paper, we introduce APSeg, an auto-prompt method for guiding SAM to complete CD-FSS tasks. To achieve fully automatic segmentation based on SAM and release the segmentation capability of SAM in cross-domain scenarios, we propose the Meta Prompt Generator (MPG) module and Dual Prototype Anchor Transformation (DPAT) module to achieve this goal. By fusing the extracted pseudo query prototypes with support prototypes, DPAT enables domain-specific input features to be more stably converted into domain-agnostic features, significantly improving cross-domain generalization capabilities. In addition, MPG generates semantic-aware prompt embeddings with meta-learning, promoting the construction of a fully automatic CD-FSS framework based on SAM. Combining DPAT and MPG, extensive experimental results show that our APSeg achieves a new state-of-the-art in CD-FSS.
\vspace{-5pt}

\section{Acknowledgements}
\vspace{-2pt}
This work is supported by the National Natural Science Foundation of China under Grant 62176163, 62306183 and 82261138629; the Science and Technology Foundation of Guangdong Province under Grant 2021A1515012303, 2024A1515010194 and 2023A1515010688; the Science and Technology Foundation of Shenzhen under Grant JCYJ20210324094602007 and JCYJ20220531101412030; and Shenzhen University Startup Funding.

{
    \small
    \bibliographystyle{ieeenat_fullname}
    \bibliography{main}

\begin{thebibliography}{11}
\providecommand{\natexlab}[1]{#1}
\providecommand{\url}[1]{\texttt{#1}}
\expandafter\ifx\csname urlstyle\endcsname\relax
  \providecommand{\doi}[1]{doi: #1}\else
  \providecommand{\doi}{doi: \begingroup \urlstyle{rm}\Url}\fi

\bibitem[Candemir et~al.(2013)Candemir, Jaeger, Palaniappan, Musco, Singh, Xue, Karargyris, Antani, Thoma, and McDonald]{candemir2013lung_SUPP}
Sema Candemir, Stefan Jaeger, Kannappan Palaniappan, Jonathan~P Musco, Rahul~K Singh, Zhiyun Xue, Alexandros Karargyris, Sameer Antani, George Thoma, and Clement~J McDonald.
\newblock Lung segmentation in chest radiographs using anatomical atlases with nonrigid registration.
\newblock \emph{IEEE transactions on medical imaging}, 33\penalty0 (2):\penalty0 577--590, 2013.

\bibitem[Cao et~al.(2022)Cao, Guo, Yuan, and Jin]{cao2022prototype_SUPP}
Leilei Cao, Yibo Guo, Ye Yuan, and Qiangguo Jin.
\newblock Prototype as query for few shot semantic segmentation.
\newblock \emph{arXiv preprint arXiv:2211.14764}, 2022.

\bibitem[Carion et~al.(2020)Carion, Massa, Synnaeve, Usunier, Kirillov, and Zagoruyko]{carion2020end_SUPP}
Nicolas Carion, Francisco Massa, Gabriel Synnaeve, Nicolas Usunier, Alexander Kirillov, and Sergey Zagoruyko.
\newblock End-to-end object detection with transformers.
\newblock In \emph{European conference on computer vision}, pages 213--229. Springer, 2020.

\bibitem[Codella et~al.(2019)Codella, Rotemberg, Tschandl, Celebi, Dusza, Gutman, Helba, Kalloo, Liopyris, Marchetti, et~al.]{codella2019skin_SUPP}
Noel Codella, Veronica Rotemberg, Philipp Tschandl, M~Emre Celebi, Stephen Dusza, David Gutman, Brian Helba, Aadi Kalloo, Konstantinos Liopyris, Michael Marchetti, et~al.
\newblock Skin lesion analysis toward melanoma detection 2018: A challenge hosted by the international skin imaging collaboration (isic).
\newblock \emph{arXiv preprint arXiv:1902.03368}, 2019.

\bibitem[Demir et~al.(2018)Demir, Koperski, Lindenbaum, Pang, Huang, Basu, Hughes, Tuia, and Raskar]{demir2018deepglobe_SUPP}
Ilke Demir, Krzysztof Koperski, David Lindenbaum, Guan Pang, Jing Huang, Saikat Basu, Forest Hughes, Devis Tuia, and Ramesh Raskar.
\newblock Deepglobe 2018: A challenge to parse the earth through satellite images.
\newblock In \emph{Proceedings of the IEEE Conference on Computer Vision and Pattern Recognition Workshops}, pages 172--181, 2018.

\bibitem[Huang et~al.(2023)Huang, Zhu, and Chen]{huang2023restnet_SUPP}
Xinyang Huang, Chuang Zhu, and Wenkai Chen.
\newblock Restnet: Boosting cross-domain few-shot segmentation with residual transformation network.
\newblock \emph{arXiv preprint arXiv:2308.13469}, 2023.

\bibitem[Jaeger et~al.(2013)Jaeger, Karargyris, Candemir, Folio, Siegelman, Callaghan, Xue, Palaniappan, Singh, Antani, et~al.]{jaeger2013automatic_SUPP}
Stefan Jaeger, Alexandros Karargyris, Sema Candemir, Les Folio, Jenifer Siegelman, Fiona Callaghan, Zhiyun Xue, Kannappan Palaniappan, Rahul~K Singh, Sameer Antani, et~al.
\newblock Automatic tuberculosis screening using chest radiographs.
\newblock \emph{IEEE transactions on medical imaging}, 33\penalty0 (2):\penalty0 233--245, 2013.

\bibitem[Lei et~al.(2022)Lei, Zhang, He, Chen, Du, and Lu]{lei2022cross_SUPP}
Shuo Lei, Xuchao Zhang, Jianfeng He, Fanglan Chen, Bowen Du, and Chang-Tien Lu.
\newblock Cross-domain few-shot semantic segmentation.
\newblock In \emph{European Conference on Computer Vision}, pages 73--90. Springer, 2022.

\bibitem[Li et~al.(2020)Li, Wei, Chen, Tai, and Tang]{li2020fss_SUPP}
Xiang Li, Tianhan Wei, Yau~Pun Chen, Yu-Wing Tai, and Chi-Keung Tang.
\newblock Fss-1000: A 1000-class dataset for few-shot segmentation.
\newblock In \emph{Proceedings of the IEEE/CVF conference on computer vision and pattern recognition}, pages 2869--2878, 2020.

\bibitem[Tian et~al.(2020)Tian, Zhao, Shu, Yang, Li, and Jia]{tian2020prior_SUPP}
Zhuotao Tian, Hengshuang Zhao, Michelle Shu, Zhicheng Yang, Ruiyu Li, and Jiaya Jia.
\newblock Prior guided feature enrichment network for few-shot segmentation.
\newblock \emph{IEEE transactions on pattern analysis and machine intelligence}, 44\penalty0 (2):\penalty0 1050--1065, 2020.

\bibitem[Zhang et~al.(2023)Zhang, Jiang, Guo, Yan, Pan, Dong, Gao, and Li]{zhang2023personalize_SUPP}
Renrui Zhang, Zhengkai Jiang, Ziyu Guo, Shilin Yan, Junting Pan, Hao Dong, Peng Gao, and Hongsheng Li.
\newblock Personalize segment anything model with one shot.
\newblock \emph{arXiv preprint arXiv:2305.03048}, 2023.

\end{thebibliography}


\begin{thebibliography}{54}
\providecommand{\natexlab}[1]{#1}
\providecommand{\url}[1]{\texttt{#1}}
\expandafter\ifx\csname urlstyle\endcsname\relax
  \providecommand{\doi}[1]{doi: #1}\else
  \providecommand{\doi}{doi: \begingroup \urlstyle{rm}\Url}\fi

\bibitem[Ben-Israel and Greville(2003)]{ben2003generalized}
Adi Ben-Israel and Thomas~NE Greville.
\newblock \emph{Generalized inverses: theory and applications}.
\newblock Springer Science \& Business Media, 2003.

\bibitem[Boudiaf et~al.(2021)Boudiaf, Kervadec, Masud, Piantanida, Ben~Ayed, and Dolz]{boudiaf2021few}
Malik Boudiaf, Hoel Kervadec, Ziko~Imtiaz Masud, Pablo Piantanida, Ismail Ben~Ayed, and Jose Dolz.
\newblock Few-shot segmentation without meta-learning: A good transductive inference is all you need?
\newblock In \emph{Proceedings of the IEEE/CVF conference on computer vision and pattern recognition}, pages 13979--13988, 2021.

\bibitem[Candemir et~al.(2013)Candemir, Jaeger, Palaniappan, Musco, Singh, Xue, Karargyris, Antani, Thoma, and McDonald]{candemir2013lung}
Sema Candemir, Stefan Jaeger, Kannappan Palaniappan, Jonathan~P Musco, Rahul~K Singh, Zhiyun Xue, Alexandros Karargyris, Sameer Antani, George Thoma, and Clement~J McDonald.
\newblock Lung segmentation in chest radiographs using anatomical atlases with nonrigid registration.
\newblock \emph{IEEE transactions on medical imaging}, 33\penalty0 (2):\penalty0 577--590, 2013.

\bibitem[Cao et~al.(2022)Cao, Guo, Yuan, and Jin]{cao2022prototype}
Leilei Cao, Yibo Guo, Ye Yuan, and Qiangguo Jin.
\newblock Prototype as query for few shot semantic segmentation.
\newblock \emph{arXiv preprint arXiv:2211.14764}, 2022.

\bibitem[Carion et~al.(2020)Carion, Massa, Synnaeve, Usunier, Kirillov, and Zagoruyko]{carion2020end}
Nicolas Carion, Francisco Massa, Gabriel Synnaeve, Nicolas Usunier, Alexander Kirillov, and Sergey Zagoruyko.
\newblock End-to-end object detection with transformers.
\newblock In \emph{European conference on computer vision}, pages 213--229. Springer, 2020.

\bibitem[Chen et~al.(2024)Chen, Liu, Chen, Zhang, Li, Zou, and Shi]{chen2024rsprompter}
Keyan Chen, Chenyang Liu, Hao Chen, Haotian Zhang, Wenyuan Li, Zhengxia Zou, and Zhenwei Shi.
\newblock Rsprompter: Learning to prompt for remote sensing instance segmentation based on visual foundation model.
\newblock \emph{IEEE Transactions on Geoscience and Remote Sensing}, 2024.

\bibitem[Chen et~al.(2017)Chen, Papandreou, Kokkinos, Murphy, and Yuille]{chen2017deeplab}
Liang-Chieh Chen, George Papandreou, Iasonas Kokkinos, Kevin Murphy, and Alan~L Yuille.
\newblock Deeplab: Semantic image segmentation with deep convolutional nets, atrous convolution, and fully connected crfs.
\newblock \emph{IEEE transactions on pattern analysis and machine intelligence}, 40\penalty0 (4):\penalty0 834--848, 2017.

\bibitem[Chen et~al.(2018)Chen, Zhu, Papandreou, Schroff, and Adam]{chen2018encoder}
Liang-Chieh Chen, Yukun Zhu, George Papandreou, Florian Schroff, and Hartwig Adam.
\newblock Encoder-decoder with atrous separable convolution for semantic image segmentation.
\newblock In \emph{Proceedings of the European conference on computer vision (ECCV)}, pages 801--818, 2018.

\bibitem[Chen et~al.(2023)Chen, Zhu, Deng, Cao, Wang, Zhang, Li, Sun, Zang, and Mao]{chen2023sam}
Tianrun Chen, Lanyun Zhu, Chaotao Deng, Runlong Cao, Yan Wang, Shangzhan Zhang, Zejian Li, Lingyun Sun, Ying Zang, and Papa Mao.
\newblock Sam-adapter: Adapting segment anything in underperformed scenes.
\newblock In \emph{Proceedings of the IEEE/CVF International Conference on Computer Vision}, pages 3367--3375, 2023.

\bibitem[Cheng et~al.(2023)Cheng, Qin, Jiang, Zhang, Lao, and Li]{cheng2023sam}
Dongjie Cheng, Ziyuan Qin, Zekun Jiang, Shaoting Zhang, Qicheng Lao, and Kang Li.
\newblock Sam on medical images: A comprehensive study on three prompt modes.
\newblock \emph{arXiv preprint arXiv:2305.00035}, 2023.

\bibitem[Codella et~al.(2019)Codella, Rotemberg, Tschandl, Celebi, Dusza, Gutman, Helba, Kalloo, Liopyris, Marchetti, et~al.]{codella2019skin}
Noel Codella, Veronica Rotemberg, Philipp Tschandl, M~Emre Celebi, Stephen Dusza, David Gutman, Brian Helba, Aadi Kalloo, Konstantinos Liopyris, Michael Marchetti, et~al.
\newblock Skin lesion analysis toward melanoma detection 2018: A challenge hosted by the international skin imaging collaboration (isic).
\newblock \emph{arXiv preprint arXiv:1902.03368}, 2019.

\bibitem[Demir et~al.(2018)Demir, Koperski, Lindenbaum, Pang, Huang, Basu, Hughes, Tuia, and Raskar]{demir2018deepglobe}
Ilke Demir, Krzysztof Koperski, David Lindenbaum, Guan Pang, Jing Huang, Saikat Basu, Forest Hughes, Devis Tuia, and Ramesh Raskar.
\newblock Deepglobe 2018: A challenge to parse the earth through satellite images.
\newblock In \emph{Proceedings of the IEEE Conference on Computer Vision and Pattern Recognition Workshops}, pages 172--181, 2018.

\bibitem[Everingham et~al.(2010)Everingham, Van~Gool, Williams, Winn, and Zisserman]{everingham2010pascal}
Mark Everingham, Luc Van~Gool, Christopher~KI Williams, John Winn, and Andrew Zisserman.
\newblock The pascal visual object classes (voc) challenge.
\newblock \emph{International journal of computer vision}, 88:\penalty0 303--338, 2010.

\bibitem[Hariharan et~al.(2011)Hariharan, Arbel{\'a}ez, Bourdev, Maji, and Malik]{hariharan2011semantic}
Bharath Hariharan, Pablo Arbel{\'a}ez, Lubomir Bourdev, Subhransu Maji, and Jitendra Malik.
\newblock Semantic contours from inverse detectors.
\newblock In \emph{2011 international conference on computer vision}, pages 991--998. IEEE, 2011.

\bibitem[He et~al.(2016)He, Zhang, Ren, and Sun]{he2016deep}
Kaiming He, Xiangyu Zhang, Shaoqing Ren, and Jian Sun.
\newblock Deep residual learning for image recognition.
\newblock In \emph{Proceedings of the IEEE conference on computer vision and pattern recognition}, pages 770--778, 2016.

\bibitem[Huang et~al.(2023)Huang, Zhu, and Chen]{huang2023restnet}
Xinyang Huang, Chuang Zhu, and Wenkai Chen.
\newblock Restnet: Boosting cross-domain few-shot segmentation with residual transformation network.
\newblock \emph{arXiv preprint arXiv:2308.13469}, 2023.

\bibitem[Jaeger et~al.(2013)Jaeger, Karargyris, Candemir, Folio, Siegelman, Callaghan, Xue, Palaniappan, Singh, Antani, et~al.]{jaeger2013automatic}
Stefan Jaeger, Alexandros Karargyris, Sema Candemir, Les Folio, Jenifer Siegelman, Fiona Callaghan, Zhiyun Xue, Kannappan Palaniappan, Rahul~K Singh, Sameer Antani, et~al.
\newblock Automatic tuberculosis screening using chest radiographs.
\newblock \emph{IEEE transactions on medical imaging}, 33\penalty0 (2):\penalty0 233--245, 2013.

\bibitem[Kingma and Ba(2014)]{kingma2014adam}
Diederik~P Kingma and Jimmy Ba.
\newblock Adam: A method for stochastic optimization.
\newblock \emph{arXiv preprint arXiv:1412.6980}, 2014.

\bibitem[Kirillov et~al.(2023)Kirillov, Mintun, Ravi, Mao, Rolland, Gustafson, Xiao, Whitehead, Berg, Lo, et~al.]{kirillov2023segment}
Alexander Kirillov, Eric Mintun, Nikhila Ravi, Hanzi Mao, Chloe Rolland, Laura Gustafson, Tete Xiao, Spencer Whitehead, Alexander~C Berg, Wan-Yen Lo, et~al.
\newblock Segment anything.
\newblock In \emph{Proceedings of the IEEE/CVF International Conference on Computer Vision}, pages 4015--4026, 2023.

\bibitem[Lang et~al.(2022)Lang, Cheng, Tu, and Han]{lang2022learning}
Chunbo Lang, Gong Cheng, Binfei Tu, and Junwei Han.
\newblock Learning what not to segment: A new perspective on few-shot segmentation.
\newblock In \emph{Proceedings of the IEEE/CVF conference on computer vision and pattern recognition}, pages 8057--8067, 2022.

\bibitem[Lei et~al.(2022)Lei, Zhang, He, Chen, Du, and Lu]{lei2022cross}
Shuo Lei, Xuchao Zhang, Jianfeng He, Fanglan Chen, Bowen Du, and Chang-Tien Lu.
\newblock Cross-domain few-shot semantic segmentation.
\newblock In \emph{European Conference on Computer Vision}, pages 73--90. Springer, 2022.

\bibitem[Li et~al.(2023)Li, Khanduri, Qiang, Sultan, Chetty, and Zhu]{li2023auto}
Chengyin Li, Prashant Khanduri, Yao Qiang, Rafi~Ibn Sultan, Indrin Chetty, and Dongxiao Zhu.
\newblock Auto-prompting sam for mobile friendly 3d medical image segmentation.
\newblock \emph{arXiv preprint arXiv:2308.14936}, 2023.

\bibitem[Li et~al.(2021)Li, Jampani, Sevilla-Lara, Sun, Kim, and Kim]{li2021adaptive}
Gen Li, Varun Jampani, Laura Sevilla-Lara, Deqing Sun, Jonghyun Kim, and Joongkyu Kim.
\newblock Adaptive prototype learning and allocation for few-shot segmentation.
\newblock In \emph{Proceedings of the IEEE/CVF conference on computer vision and pattern recognition}, pages 8334--8343, 2021.

\bibitem[Li et~al.(2020)Li, Wei, Chen, Tai, and Tang]{li2020fss}
Xiang Li, Tianhan Wei, Yau~Pun Chen, Yu-Wing Tai, and Chi-Keung Tang.
\newblock Fss-1000: A 1000-class dataset for few-shot segmentation.
\newblock In \emph{Proceedings of the IEEE/CVF conference on computer vision and pattern recognition}, pages 2869--2878, 2020.

\bibitem[Long et~al.(2015)Long, Shelhamer, and Darrell]{long2015fully}
Jonathan Long, Evan Shelhamer, and Trevor Darrell.
\newblock Fully convolutional networks for semantic segmentation.
\newblock In \emph{Proceedings of the IEEE conference on computer vision and pattern recognition}, pages 3431--3440, 2015.

\bibitem[Lu et~al.(2021)Lu, He, Zhu, Zhang, Song, and Xiang]{lu2021simpler}
Zhihe Lu, Sen He, Xiatian Zhu, Li Zhang, Yi-Zhe Song, and Tao Xiang.
\newblock Simpler is better: Few-shot semantic segmentation with classifier weight transformer.
\newblock In \emph{Proceedings of the IEEE/CVF International Conference on Computer Vision}, pages 8741--8750, 2021.

\bibitem[Mazurowski et~al.(2023)Mazurowski, Dong, Gu, Yang, Konz, and Zhang]{mazurowski2023segment}
Maciej~A Mazurowski, Haoyu Dong, Hanxue Gu, Jichen Yang, Nicholas Konz, and Yixin Zhang.
\newblock Segment anything model for medical image analysis: an experimental study.
\newblock \emph{Medical Image Analysis}, 89:\penalty0 102918, 2023.

\bibitem[Min et~al.(2021)Min, Kang, and Cho]{min2021hypercorrelation}
Juhong Min, Dahyun Kang, and Minsu Cho.
\newblock Hypercorrelation squeeze for few-shot segmentation.
\newblock In \emph{Proceedings of the IEEE/CVF international conference on computer vision}, pages 6941--6952, 2021.

\bibitem[Oquab et~al.(2023)Oquab, Darcet, Moutakanni, Vo, Szafraniec, Khalidov, Fernandez, Haziza, Massa, El-Nouby, et~al.]{oquab2023dinov2}
Maxime Oquab, Timoth{\'e}e Darcet, Th{\'e}o Moutakanni, Huy Vo, Marc Szafraniec, Vasil Khalidov, Pierre Fernandez, Daniel Haziza, Francisco Massa, Alaaeldin El-Nouby, et~al.
\newblock Dinov2: Learning robust visual features without supervision.
\newblock \emph{arXiv preprint arXiv:2304.07193}, 2023.

\bibitem[Peng et~al.(2023)Peng, Tian, Wu, Wang, Liu, Su, and Jia]{peng2023hierarchical}
Bohao Peng, Zhuotao Tian, Xiaoyang Wu, Chengyao Wang, Shu Liu, Jingyong Su, and Jiaya Jia.
\newblock Hierarchical dense correlation distillation for few-shot segmentation.
\newblock In \emph{Proceedings of the IEEE/CVF Conference on Computer Vision and Pattern Recognition}, pages 23641--23651, 2023.

\bibitem[Radford et~al.(2021)Radford, Kim, Hallacy, Ramesh, Goh, Agarwal, Sastry, Askell, Mishkin, Clark, et~al.]{radford2021learning}
Alec Radford, Jong~Wook Kim, Chris Hallacy, Aditya Ramesh, Gabriel Goh, Sandhini Agarwal, Girish Sastry, Amanda Askell, Pamela Mishkin, Jack Clark, et~al.
\newblock Learning transferable visual models from natural language supervision.
\newblock In \emph{International conference on machine learning}, pages 8748--8763. PMLR, 2021.

\bibitem[Russakovsky et~al.(2015)Russakovsky, Deng, Su, Krause, Satheesh, Ma, Huang, Karpathy, Khosla, Bernstein, et~al.]{russakovsky2015imagenet}
Olga Russakovsky, Jia Deng, Hao Su, Jonathan Krause, Sanjeev Satheesh, Sean Ma, Zhiheng Huang, Andrej Karpathy, Aditya Khosla, Michael Bernstein, et~al.
\newblock Imagenet large scale visual recognition challenge.
\newblock \emph{International journal of computer vision}, 115:\penalty0 211--252, 2015.

\bibitem[Shaban et~al.(2017)Shaban, Bansal, Liu, Essa, and Boots]{shaban2017one}
Amirreza Shaban, Shray Bansal, Zhen Liu, Irfan Essa, and Byron Boots.
\newblock One-shot learning for semantic segmentation.
\newblock \emph{arXiv preprint arXiv:1709.03410}, 2017.

\bibitem[Siam et~al.(2019)Siam, Oreshkin, and Jagersand]{siam2019amp}
Mennatullah Siam, Boris~N Oreshkin, and Martin Jagersand.
\newblock Amp: Adaptive masked proxies for few-shot segmentation.
\newblock In \emph{Proceedings of the IEEE/CVF International Conference on Computer Vision}, pages 5249--5258, 2019.

\bibitem[Simonyan and Zisserman(2014)]{simonyan2014very}
Karen Simonyan and Andrew Zisserman.
\newblock Very deep convolutional networks for large-scale image recognition.
\newblock \emph{arXiv preprint arXiv:1409.1556}, 2014.

\bibitem[Snell et~al.(2017)Snell, Swersky, and Zemel]{snell2017prototypical}
Jake Snell, Kevin Swersky, and Richard Zemel.
\newblock Prototypical networks for few-shot learning.
\newblock \emph{Advances in neural information processing systems}, 30, 2017.

\bibitem[Tavera et~al.(2022)Tavera, Cermelli, Masone, and Caputo]{tavera2022pixel}
Antonio Tavera, Fabio Cermelli, Carlo Masone, and Barbara Caputo.
\newblock Pixel-by-pixel cross-domain alignment for few-shot semantic segmentation.
\newblock In \emph{Proceedings of the IEEE/CVF Winter Conference on Applications of Computer Vision}, pages 1626--1635, 2022.

\bibitem[Tian et~al.(2020)Tian, Zhao, Shu, Yang, Li, and Jia]{tian2020prior}
Zhuotao Tian, Hengshuang Zhao, Michelle Shu, Zhicheng Yang, Ruiyu Li, and Jiaya Jia.
\newblock Prior guided feature enrichment network for few-shot segmentation.
\newblock \emph{IEEE transactions on pattern analysis and machine intelligence}, 44\penalty0 (2):\penalty0 1050--1065, 2020.

\bibitem[Wald et~al.(2023)Wald, Roy, Koehler, Disch, Rokuss, Holzschuh, Zimmerer, and Maier-Hein]{roy2023sam}
Tassilo Wald, Saikat Roy, Gregor Koehler, Nico Disch, Maximilian~Rouven Rokuss, Julius Holzschuh, David Zimmerer, and Klaus Maier-Hein.
\newblock Sam. md: Zero-shot medical image segmentation capabilities of the segment anything model.
\newblock In \emph{Medical Imaging with Deep Learning, short paper track}, 2023.

\bibitem[Wang et~al.(2023)Wang, Islam, Xu, Zhang, and Ren]{wang2023sam}
An Wang, Mobarakol Islam, Mengya Xu, Yang Zhang, and Hongliang Ren.
\newblock Sam meets robotic surgery: An empirical study on generalization, robustness and adaptation.
\newblock In \emph{International Conference on Medical Image Computing and Computer-Assisted Intervention}, pages 234--244. Springer, 2023.

\bibitem[Wang et~al.(2019)Wang, Liew, Zou, Zhou, and Feng]{wang2019panet}
Kaixin Wang, Jun~Hao Liew, Yingtian Zou, Daquan Zhou, and Jiashi Feng.
\newblock Panet: Few-shot image semantic segmentation with prototype alignment.
\newblock In \emph{proceedings of the IEEE/CVF international conference on computer vision}, pages 9197--9206, 2019.

\bibitem[Wang et~al.(2022{\natexlab{a}})Wang, Duan, Wang, En, Fan, and Zhang]{wang2022remember}
Wenjian Wang, Lijuan Duan, Yuxi Wang, Qing En, Junsong Fan, and Zhaoxiang Zhang.
\newblock Remember the difference: Cross-domain few-shot semantic segmentation via meta-memory transfer.
\newblock In \emph{Proceedings of the IEEE/CVF Conference on Computer Vision and Pattern Recognition}, pages 7065--7074, 2022{\natexlab{a}}.

\bibitem[Wang et~al.(2022{\natexlab{b}})Wang, Xu, Tian, Luo, Shi, Zhang, Fan, and He]{wang2022cross}
Yixin Wang, Zhe Xu, Jiang Tian, Jie Luo, Zhongchao Shi, Yang Zhang, Jianping Fan, and Zhiqiang He.
\newblock Cross-domain few-shot learning for rare-disease skin lesion segmentation.
\newblock In \emph{ICASSP 2022-2022 IEEE International Conference on Acoustics, Speech and Signal Processing (ICASSP)}, pages 1086--1090. IEEE, 2022{\natexlab{b}}.

\bibitem[Wu et~al.(2021)Wu, Shi, Lin, and Cai]{wu2021learning}
Zhonghua Wu, Xiangxi Shi, Guosheng Lin, and Jianfei Cai.
\newblock Learning meta-class memory for few-shot semantic segmentation.
\newblock In \emph{Proceedings of the IEEE/CVF International Conference on Computer Vision}, pages 517--526, 2021.

\bibitem[Xie et~al.(2021)Xie, Wang, Yu, Anandkumar, Alvarez, and Luo]{xie2021segformer}
Enze Xie, Wenhai Wang, Zhiding Yu, Anima Anandkumar, Jose~M Alvarez, and Ping Luo.
\newblock Segformer: Simple and efficient design for semantic segmentation with transformers.
\newblock \emph{Advances in Neural Information Processing Systems}, 34:\penalty0 12077--12090, 2021.

\bibitem[Yang et~al.(2020)Yang, Liu, Li, Jiao, and Ye]{yang2020prototype}
Boyu Yang, Chang Liu, Bohao Li, Jianbin Jiao, and Qixiang Ye.
\newblock Prototype mixture models for few-shot semantic segmentation.
\newblock In \emph{Computer Vision--ECCV 2020: 16th European Conference, Glasgow, UK, August 23--28, 2020, Proceedings, Part VIII 16}, pages 763--778. Springer, 2020.

\bibitem[Yang et~al.(2021)Yang, Zhuo, Qi, Shi, and Gao]{yang2021mining}
Lihe Yang, Wei Zhuo, Lei Qi, Yinghuan Shi, and Yang Gao.
\newblock Mining latent classes for few-shot segmentation.
\newblock In \emph{Proceedings of the IEEE/CVF international conference on computer vision}, pages 8721--8730, 2021.

\bibitem[Yue et~al.(2023)Yue, Zhang, Hu, Xia, Luo, and Wang]{yue2023surgicalsam}
Wenxi Yue, Jing Zhang, Kun Hu, Yong Xia, Jiebo Luo, and Zhiyong Wang.
\newblock Surgicalsam: Efficient class promptable surgical instrument segmentation.
\newblock \emph{arXiv preprint arXiv:2308.08746}, 2023.

\bibitem[Zhang et~al.(2021{\natexlab{a}})Zhang, Xiao, and Qin]{zhang2021self}
Bingfeng Zhang, Jimin Xiao, and Terry Qin.
\newblock Self-guided and cross-guided learning for few-shot segmentation.
\newblock In \emph{Proceedings of the IEEE/CVF Conference on Computer Vision and Pattern Recognition}, pages 8312--8321, 2021{\natexlab{a}}.

\bibitem[Zhang et~al.(2019{\natexlab{a}})Zhang, Lin, Liu, Guo, Wu, and Yao]{zhang2019pyramid}
Chi Zhang, Guosheng Lin, Fayao Liu, Jiushuang Guo, Qingyao Wu, and Rui Yao.
\newblock Pyramid graph networks with connection attentions for region-based one-shot semantic segmentation.
\newblock In \emph{Proceedings of the IEEE/CVF International Conference on Computer Vision}, pages 9587--9595, 2019{\natexlab{a}}.

\bibitem[Zhang et~al.(2019{\natexlab{b}})Zhang, Lin, Liu, Yao, and Shen]{zhang2019canet}
Chi Zhang, Guosheng Lin, Fayao Liu, Rui Yao, and Chunhua Shen.
\newblock Canet: Class-agnostic segmentation networks with iterative refinement and attentive few-shot learning.
\newblock In \emph{Proceedings of the IEEE/CVF conference on computer vision and pattern recognition}, pages 5217--5226, 2019{\natexlab{b}}.

\bibitem[Zhang et~al.(2021{\natexlab{b}})Zhang, Kang, Yang, and Wei]{zhang2021few}
Gengwei Zhang, Guoliang Kang, Yi Yang, and Yunchao Wei.
\newblock Few-shot segmentation via cycle-consistent transformer.
\newblock \emph{Advances in Neural Information Processing Systems}, 34:\penalty0 21984--21996, 2021{\natexlab{b}}.

\bibitem[Zhang et~al.(2023)Zhang, Jiang, Guo, Yan, Pan, Dong, Gao, and Li]{zhang2023personalize}
Renrui Zhang, Zhengkai Jiang, Ziyu Guo, Shilin Yan, Junting Pan, Hao Dong, Peng Gao, and Hongsheng Li.
\newblock Personalize segment anything model with one shot.
\newblock \emph{arXiv preprint arXiv:2305.03048}, 2023.

\bibitem[Zhang et~al.(2020)Zhang, Wei, Yang, and Huang]{zhang2020sg}
Xiaolin Zhang, Yunchao Wei, Yi Yang, and Thomas~S Huang.
\newblock Sg-one: Similarity guidance network for one-shot semantic segmentation.
\newblock \emph{IEEE transactions on cybernetics}, 50\penalty0 (9):\penalty0 3855--3865, 2020.

\end{thebibliography}
}

\clearpage
\setcounter{page}{1}
\maketitlesupplementary

\setcounter{section}{0}
\setcounter{section}{0}
\section{Experiment}
\label{sec:rationale}
\subsection{Datasets}
\noindent\textbf{FSS-1000.} FSS-1000 \citeExtra{li2020fss_SUPP} is a natural scenario dataset containing 1000 class categories with 10 samples per category. The evaluation procedure is conducted on 2400 randomly sampled support-query pairs.

\noindent\textbf{Chest X-ray.} Chest X-ray \citeExtra{candemir2013lung_SUPP, jaeger2013automatic_SUPP} is an X-ray image dataset of 566 images collected from 58 cases with manifestations of tuberculosis and 80 normal cases.

\noindent\textbf{ISIC.} ISIC \citeExtra{codella2019skin_SUPP} is a skin lesion image dataset from the ISIC-2018 challenge. Following the previous approach \citeExtra{lei2022cross_SUPP}, the evaluation procedure is conducted on the training set, which includes 2596 images and the corresponding annotations.

\noindent\textbf{Deepglobe.} Deepglobe \citeExtra{demir2018deepglobe_SUPP} is a remote sensing image dataset that can be used for land cover segmentation. The dataset contains 7 categories: areas of urban, agriculture, rangeland, forest, water, barrel, and unknown. Following the previous method \citeExtra{lei2022cross_SUPP}, we filter the unknown class in the training set and chunk the images to obtain 5666 images. We report the test results on the processed training set.
\subsection{Implementation Details} 
In the meta prompt generator (MPG) module, the spatial size of the feature enrichment module (FEM) is set to $\left\{60, 30, 15, 8\right\}$, maintaining consistency with PFENet \citeExtra{tian2020prior_SUPP}. The transformer decoder~\citeExtra{carion2020end_SUPP} block consists of a self-attention mechanism, a cross-attention mechanism, and a feed-forward network. Its configuration is in line with Protoformer \citeExtra{cao2022prototype_SUPP}.

\subsection{Ablation Study} 
\noindent\textbf{Effect of the cycle consistent selection.} In the dual prototype anchor transformation (DPAT) module, pseudo query prototypes are extracted through cycle-consistent selection (CSS) to enhance the feature transformation process. To further validate the effectiveness of CCS, we explore an alternative method for extracting pseudo query prototypes. Analogous to ResNet \citeExtra{huang2023restnet_SUPP}, we first obtain the coarse prediction mask of the query image and then perform the MAP operation to obtain query prototypes. This method is referred to as prediction mask-based MAP (PM-MAP). We conduct an experiment to evaluate the model without CCS, with CCS, and with PM-MAP, respectively, to better analyze the contribution of our CCS. The results in Tab.~\ref{supp_table1} show that CCS achieves better performance, with an average improvement of 2.67\% mIoU on four datasets on the 1-shot setting. This indicates that CCS can extract reliable query prototypes to enhance the support prototypes, allowing features to be transformed into a more stable domain-agnostic space. Qualitative results on the effectiveness of CSS are provided in Fig.~\ref{fig:supp-1}.

\setcounter{table}{0}
\begin{table}
\centering
\setlength{\tabcolsep}{7mm}{
    \scalebox{0.73}{
        \begin{tabular}{c|c}
            \toprule
            Method &1-shot mIoU \\
            \midrule
            w/o CCS       & 58.63   \\
            w/ CCS        & 61.30         \\
            w/ PM-MAP     & 59.05         \\
            \bottomrule
        \end{tabular}
    }
}
\caption{Ablation studies of cycle-consistent selection (CSS) in dual prototype anchor transformation (DPAT) module. PM-MAP means the prediction mask-based MAP method. The results are averaged over four datasets under the 1-shot setting.}
\label{supp_table1}
\vspace{-10pt}
\end{table}

\subsection{Additional Analysis} 
PerSAM \citeExtra{zhang2023personalize_SUPP} is a few-shot segmentation method based on SAM. To enable automatic segmentation, PerSAM extracts point prompts based on cosine similarity measure and obtains box/mask prompts from the coarse predictions. The final prediction results are produced under the guidance of three types of visual prompts. The performance of PerSAM is far inferior to our proposed APSeg and PATNet~\citeExtra{lei2022cross_SUPP} due to the inability to extract precise prompts. In contrast, our APSeg introduces MPG and DPAT, avoiding reliance on precise visual prompts and achieving competitive results in cross-domain scenarios. Qualitative results are provided in Fig.~\ref{fig:supp-2}. It can be observed that our method outperforms PerSAM by a large margin, which validates the effectiveness of our automatic way of generating prompt embeddings. In addition, we provide more qualitative segmentation results of our proposed method on four datasets in Fig.~\ref{fig:supp-3}.

{
    \small
    \bibliographystyleExtra{ieeenat_fullname}
    \bibliographyExtra{supp}
}

\setcounter{figure}{0}
\begin{figure*}[htbp]
\centering
\includegraphics[width=0.9\linewidth]{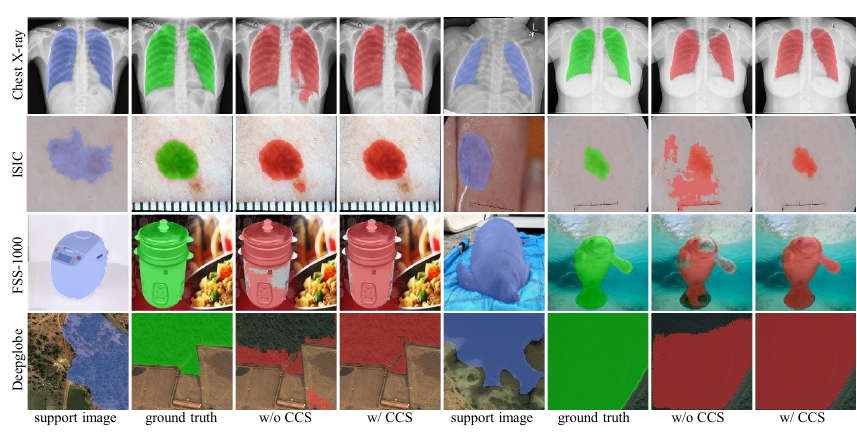}
\caption{Visual comparison of segmentation results with and without cycle-consistent selection (CCS) in dual prototype anchor transformation (DPAT) module under the 1-shot setting.}
\label{fig:supp-1}
\end{figure*}

\begin{figure*}[htbp]
\centering
\includegraphics[width=0.9\linewidth]{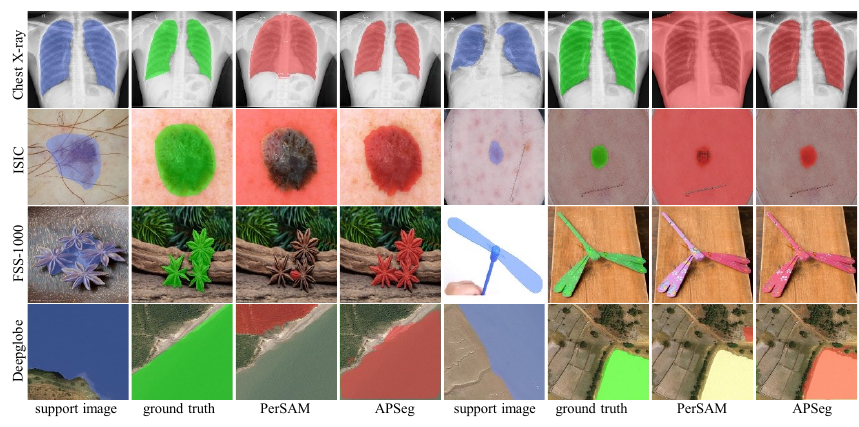}
\caption{Visual Comparison Results between APSeg and PerSAM in four target datasets under the 1-shot setting.}
\label{fig:supp-2}
\end{figure*}

\begin{figure*}[htbp]
\centering
\includegraphics[width=0.9\linewidth]{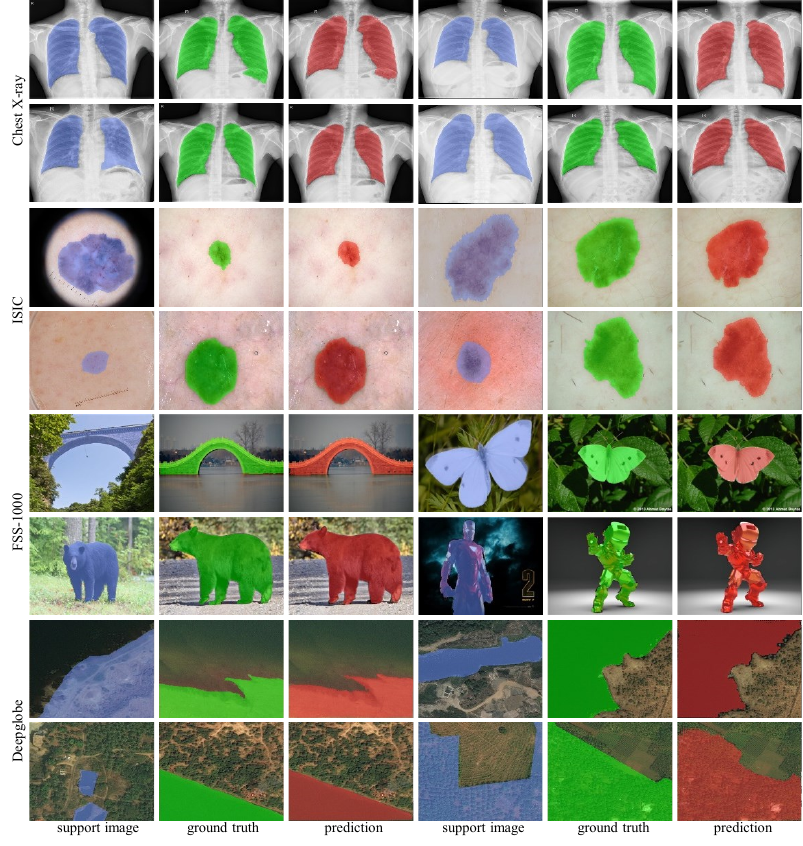}
\caption{More qualitative segmentation results of our proposed APSeg in four target datasets under the 1-shot setting.}
\label{fig:supp-3}
\end{figure*}
\end{document}